\documentclass[a4paper,conference]{IEEEtran}
\IEEEoverridecommandlockouts
\usepackage{amsmath,amssymb,amsfonts}
\usepackage[ruled,vlined]{algorithm2e}
\usepackage{booktabs}
\usepackage{caption}
\usepackage{capt-of}
\usepackage{comment}
\usepackage{color}
\usepackage{float}
\usepackage{graphicx}
\usepackage[pdftex]{hyperref}
\usepackage{multirow}
\usepackage{nameref}
\usepackage{subfigure}
\usepackage{tabularx}
\usepackage{textcomp}
\usepackage[noadjust]{cite}
\def\BibTeX{{\rm B\kern-.05em{\sc i\kern-.025em b}\kern-.08em
		T\kern-.1667em\lower.7ex\hbox{E}\kern-.125emX}}

\DeclareMathOperator*{\argmin}{arg\,min}

\begin{document}
	
	\title{Meta Soft Label Generation for Noisy Labels}
	
	\author{
		\IEEEauthorblockN{G\"{o}rkem Algan\IEEEauthorrefmark{1}\IEEEauthorrefmark{2}, Ilkay Ulusoy\IEEEauthorrefmark{1}}
		\IEEEauthorblockA{\IEEEauthorrefmark{1}Middle East Technical University, Electrical-Electronics Engineering
			\\\{e162565, ilkay\}@metu.edu.tr}
		\IEEEauthorblockA{\IEEEauthorrefmark{2}ASELSAN
			\\galgan@aselsan.com.tr}
	}
	
	\maketitle
	
	\begin{abstract}
		The existence of noisy labels in the dataset causes significant performance degradation for deep neural networks (DNNs). To address this problem, we propose a Meta Soft Label Generation algorithm called MSLG, which can jointly generate soft labels using meta-learning techniques and learn DNN parameters in an end-to-end fashion. Our approach adapts the meta-learning paradigm to estimate optimal label distribution by checking gradient directions on both noisy training data and noise-free meta-data. In order to iteratively update soft labels, meta-gradient descent step is performed on estimated labels, which would minimize the loss of noise-free meta samples. In each iteration, the base classifier is trained on estimated meta labels. MSLG is model-agnostic and can be added on top of any existing model at hand with ease. We performed extensive experiments on CIFAR10, Clothing1M and Food101N datasets. Results show that our approach outperforms other state-of-the-art methods by a large margin. Our code is available at \url{https://github.com/gorkemalgan/MSLG_noisy_label}.
	\end{abstract}
	
	\begin{IEEEkeywords}
		deep learning, label noise, noise robust, noise cleansing, meta-learning
	\end{IEEEkeywords}

	\section{Introduction} \label{introduction}
Recent advancements in deep learning have led to great improvements in computer vision systems \cite{krizhevsky2012imagenet,he2016deep,simonyan2014very}. Even though it is shown that deep networks have an impressive ability to generalize \cite{rolnick2017deep}, these powerful models have a high tendency to memorize even complete random noise \cite{zhang2016understanding,krueger2017deep,arpit2017closer}. Therefore, avoiding memorization is an essential challenge to overcome to obtain representative neural networks, and it gets even more crucial in the presence of noise. There are two types of noises, namely: feature noise and label noise \cite{frenay2014classification}. Generally speaking, label noise is considered to be more harmful than feature noise \cite{zhu2004class}.

Gathering cleanly annotated large datasets are both time consuming and expensive. Furthermore, in some fields, where labeling requires a certain amount of expertise, such as medical imaging, even experts may have contradicting opinions about labels, which would result in label noise \cite{guan2018said}. Therefore, datasets used in practical applications mostly contain noisy labels. Due to its common occurrence, research on deep learning techniques in the presence of noisy labels has gained popularity, and there are various approaches proposed in the literature \cite{algan2019image}. 

In this work, we propose a soft-label generation framework that simultaneously seeks optimal label distribution and network parameters. 

Each iteration of the training loop consists of two stages. In the first stage, posterior label distribution is updated by gradient descent step on the meta objective. Our meta objective based on a basic assumption: \textit{the best training label distribution should minimize the loss of the small clean meta-data}. Therefore, the meta-data is chosen as the clean subset of the dataset. In the second stage, the base classifier is trained on images with their corresponding predicted labels. Our framework simultaneously generates soft-label posterior distribution and learns network parameters, by consecutively repeating these two stages.

Our contribution to the literature can be summarized as follows:
\begin{itemize}
    \item We propose an end-to-end framework called MSLG, which can simultaneously generate meta-soft-labels and learn base classifier parameters. It seeks for label distribution that would maximize gradient steps on meta dataset by checking the consistency of gradient directions.
    \item Our algorithm needs a clean subset of data as meta-data. However, required clean subset of data is very small compared to whole dataset (less than 2\% of training data), which can easily be obtained in many cases. Our algorithm can effectively eliminate noise even in the extreme scenarios such as 80\% noise.
    \item Our proposed solution is model agnostic and can easily be adopted by any classification architecture at hand. It requires minimal hyper-parameter tuning and can easily adapt to various datasets from various domains.
    \item We conduct extensive experiments to show the effectiveness of the proposed method under both synthetic and real-world noise scenarios.
\end{itemize}

This paper is organized as follows. In \autoref{relatedwork}, we present related works from the literature. The proposed approach is explained in \autoref{method}, and experimental results are provided in \autoref{experiments}. Finally, \autoref{conclusion} concludes the paper by further discussion on the topic.	
	\section{Related Work} \label{relatedwork}
In this section, we first present various approaches from the literature. Then we focus on fields of label noise cleansing and meta-learning, which this work belongs to.

\textbf{Various approaches in the literature:} One line of works tries to determine the noise transition matrix \cite{sukhbaatar2014training,sukhbaatar2014learning,bekker2016training,patrini2017making}, which is the probabilistic mapping from true labels to noisy labels. Then, this matrix is used in loss function for \textit{loss correction}. However, matrix size increases exponentially with an increasing number of classes, which makes the problem intractable for large scale datasets. Moreover, these methods assume that noise is class-dependent, which is not a valid assumption for more complicated noises such as feature-dependent noises. Another line of works focuses on robust losses \cite{ghosh2017robust}. Works show that 0-1 loss has more noise tolerance than traditional loss functions \cite{manwani2013noise,ghosh2015making,charoenphakdee2019symmetric}, but it is not feasible to use it in learning since it is a non-convex loss function. Therefore, various alternative loss functions for noisy data are proposed such as; mean absolute value of error \cite{ghosh2017robust}, improved mean absolute value of error \cite{wang2019improved}, generalized cross-entropy \cite{zhang2018generalized}, symmetric cross entropy \cite{wang2019symmetric}. These methods generally rely on the internal noise robustness of DNNs and try to enhance this ability with proposed robust losses. However, they do not consider the fact that DNNs can learn from uninformative random data \cite{zhang2016understanding}. Regularizer based approach is proposed in \cite{ma2018dimensionality}, defending performance degradation is due to overfitting of noisy data. Some works aim to emphasize likely to be clean samples during training. There are two major lines of approaches for this kind of methods. The first approach is to rank samples according to their cleanness, and then samples are fed to the network by a curriculum going from clean samples to noisy samples \cite{jiang2017mentornet,han2018progressive,reed2014training,han2018co}. The second approach uses a weighting factor for each sample so that their impact on learning is weighted according to their reliability \cite{wang2018iterative,lee2018cleannet}. Please see \cite{algan2019image} for a more detailed survey on the field.

\textbf{Label noise cleansing algorithms:} One obvious way to clean noise is to remove suspicious samples \cite{northcutt2017learning} or their labels \cite{ding2018semi} from the dataset. Nevertheless, this results in an undesirable loss of information. Therefore, works mainly focus on correcting noisy labels iteratively during training. A joint optimization framework for both training base classifier and propagating noisy labels to cleaner labels is presented in \cite{tanaka2018joint}. Using expectation-maximization, both classifier parameters and label posterior distribution is estimated in order to minimize the loss. A similar approach is used in \cite{yi2019probabilistic} with additional compatibility loss condition on label posterior. Considering noisy labels are in the minority, this term assures posterior label distribution is not diverged too much from given noisy label distribution, so that majority of the clean label contribution is not lost. Our proposed solution is slightly different than label cleansing algorithms. Instead of finding exact clean label distribution, our framework aims to find optimal soft-label distribution that would maximize the learning on the small noise-free data. We call these estimated labels \textit{meta-soft-labels}.

\textbf{Meta learning algorithms:} As a meta-learning paradigm, MAML \cite{finn2017model} shown to give fruitful results in many fields. MAML aims to take gradient steps that are consistent with meta objective. This idea is also applied to noisy label setup. For example, \cite{junnan2018learning} aims to find most noise-tolerant weight initialization for the base network. Three particular works worth mentioning are \cite{ren2018learning,jenni2018deep,shu2019meta}, in which authors try to find the best sample weighting scheme for noise robustness. Their meta objective is defined to minimize the loss on the noise-free subset of data. Our approach is similar to these works, but our objective is totally different. Instead of searching for a weighting scheme, our framework seeks for a soft-label distribution.

	\section{The Proposed MSLG Method} \label{method}

\begin{figure*}[h]
	\centering
	\includegraphics[width=\textwidth]{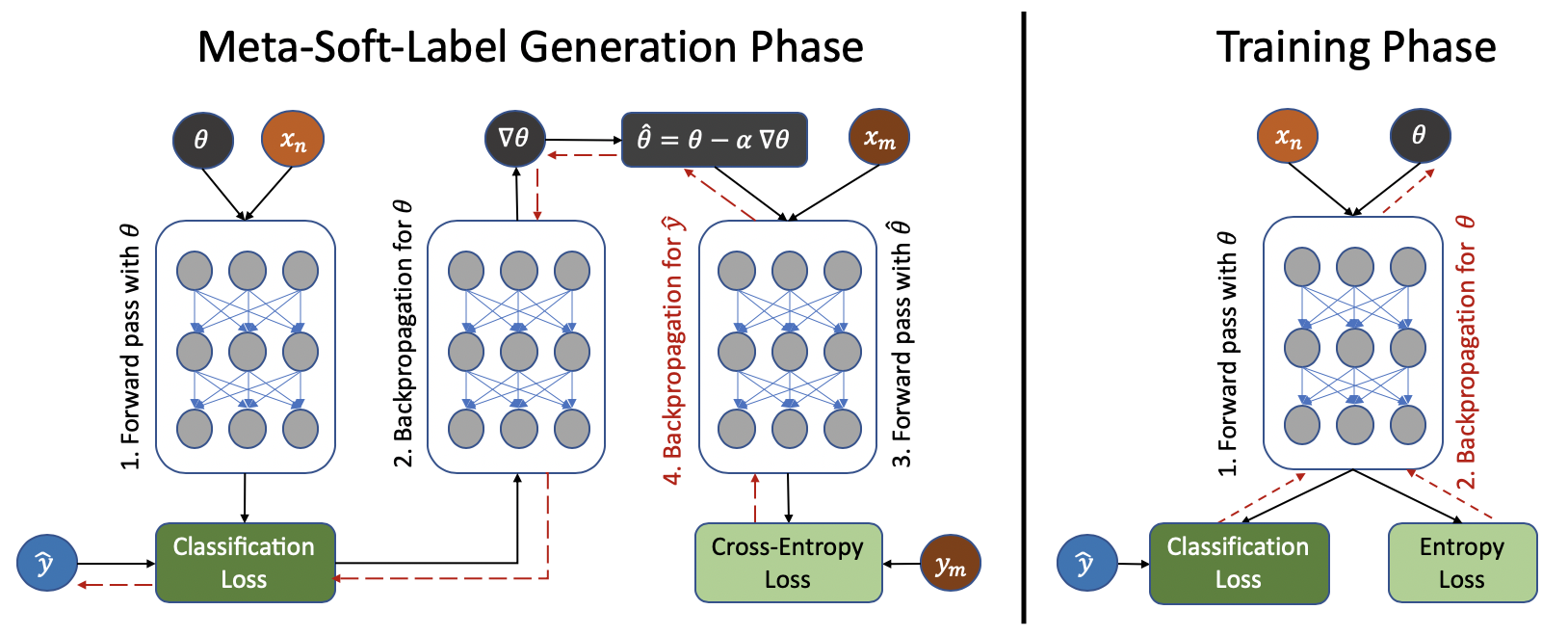}
	\caption{Training consists of two consecutive stages. In the first stage, $\theta$ is updated on noisy training data $x_n$ and its predicted label $\hat{y}$. Then forward pass is done with updated parameters $\hat{\theta}$ on meta data $x_m$ with clean labels $y_m$. Afterward, gradients are backpropagated for optimal prediction of $\hat{y}$. In the second stage, network is trained on noisy data batch with predicted labels and additional entropy loss.}
	\label{fig:mlnc}
\end{figure*} 

\subsection{Problem Statement}
Classical supervised learning consists of an input dataset $\mathcal{S}=\{(x_1,y_1),...,(x_N,y_N)\}\in (X,Y)^N$ drawn according to an unknown distribution $\mathcal{D}$, over $(X,Y)$. Task is to find the best mapping function $f:X \rightarrow Y$ among family of functions $\mathcal{F}$, where each function is parametrized by $\theta$.

One way of evaluating the performance of a classifier is the so called loss function, denoted as $l$. Given an example $(x_i,y_i) \in (X,Y)$, $l(f_\theta(x_i),y_i)$ evaluates how good is the classifier prediction. Then, for any classifier $f_\theta$, the expected risk is defined as follows

\begin{equation}
    R_{l,\mathcal{D}}(f_\theta)=E_\mathcal{D}[l(f_\theta(x),y)]
\end{equation}

where E denotes the expectation over distribution $\mathcal{D}$. Since it is not generally feasible to have complete knowledge over distribution $\mathcal{D}$, as an approximation, the empirical risk is used

\begin{equation}
    \hat{R}_{l,\mathcal{D}}(f_\theta)=\dfrac{1}{N}\sum_{i=1}^{N}l(f_\theta(x_i),y_i)
\end{equation}

Various methods of learning a classifier may be seen as minimizing the empirical risk subjected to network parameters 

\begin{equation}
    \theta^\star = \underset{\theta}{\argmin}\hat{R}_{l,\mathcal{D}}(f_\theta)
\end{equation}

In the presence of the noise, dataset turns into $\mathcal{S}_n=\{(x_1,\tilde{y}_1),...,(x_N,\tilde{y}_N)\}\in (X,Y)^N$ drawn according to a noisy distribution $\mathcal{D}_n$, over $(X,Y)$. Then, risk minimization results in 

\begin{equation}
    \theta_n^\star = \underset{\theta}{\argmin}\hat{R}_{l,\mathcal{D}_n}(f_\theta)
\end{equation}

As a result, obtained parameters by minimizing over $\mathcal{D}_n$ are different from desired optimal classifier parameters

\begin{center}
    $\theta^\star \neq \theta_n^\star$
\end{center}

Therefore, our proposed framework seeks for optimal label distribution $\hat{\mathcal{D}}$ in order to train optimal network parameters $\theta^\star$ for distribution $\mathcal{D}$. Throughout this paper, we represent noise-free labels by $y$, noisy labels by $\tilde{y}$, and predicted labels by $\hat{y}$. While $y,\tilde{y}$ are hard labels, $\hat{y}$ is soft label. $\mathcal{D}_n$ represents the noisy training data distribution and $\mathcal{D}_m$ represents the clean meta-data distribution. $N$ is the number of training data samples and $M$ is the number of meta-data samples, where $M<<N$. We represent the number of classes as $C$ and superscript represents the label probability for that class, such that $y_i^j$ represents the label value of $i^{th}$ sample for class $j$.

\subsection{Learning with MSLG}
The overall architecture of MSLG is illustrated in \autoref{fig:mlnc}, which consists of two consecutive stages: meta-soft-label generation correction and training. For each batch of data, first posterior label distribution is updated according to meta-objective, and then the base classifier is trained on these predicted labels. 

\subsubsection{\textbf{Meta-Soft-Label Generation}} This phase consists of two major steps. First, we find updated network parameters by using mini-batch from training samples and corresponding predicted labels. This is done by taking a stochastic gradient descent (SGD) step on classification loss

\begin{equation}
    \hat{\theta} = \theta^{(t)} - \alpha \dfrac{1}{N} \sum_{i=1}^{N} \nabla_\theta \mathcal{L}^{n}_i(\theta,\hat{y}_i^{(t)})\Biggr|_{\theta^{(t)}}
    \label{eq:thetahat}
\end{equation}

where $\mathcal{L}^{n}_i(\theta,\hat{y}_i^{(t)}) = \mathcal{L}_c(f_\theta(x_i),\hat{y}_i)$ for $x_i \in \mathcal{D}_n$ and $\hat{y}_i^{(t)}$ is corresponding predicted soft label at time step $t$. $\mathcal{L}_c$ represents the classification loss, which is further explained in \autoref{lossfunctions}. Secondly, we update label predictions by minimizing the loss on meta-data with the feedback coming from updated parameters.

\begin{equation}
    \hat{y}^{(t+1)} = \hat{y}^{(t)} - \beta \dfrac{1}{M} \sum_{i=1}^{M} \nabla_{\hat{y}} \mathcal{L}^{m}_i(\hat{\theta}, y_i)\Biggr|_{\hat{y}^{(t)}}
    \label{eq:ymeta}
\end{equation}

where $\mathcal{L}^{m}_i(\hat{\theta}) = \mathcal{L}_{cce}(f_{\hat{\theta}}(x_i),y_i)$ for $x_i,y_i \in \mathcal{D}_m$ and $\mathcal{L}_{cce}$ represents the classical \textit{categorical cross entropy loss}. 

\subsubsection{\textbf{Training}} In this phase, network parameters are updated with SGD on noisy training samples with corrected label predictions and entropy loss

\begin{equation}
    \theta^{(t+1)} = \theta^{(t)} - \lambda \dfrac{1}{N} \sum_{i=1}^{N} \nabla_\theta \left(\mathcal{L}_{c}(\theta,\hat{y}_i^{(t+1)})\Biggr|_{\theta^{(t)}} + \mathcal{L}_{e}(f_\theta(x_i))\right)
    \label{eq:train}
\end{equation}

Notice that we have three different learning rates $\alpha, \beta$ and $\lambda$ for each step. The overall algorithm is summarized in \autoref{algo}.

\begin{algorithm}
    \SetAlgoLined
    \KwIn{Training data $\mathcal{D}_n$, meta-data $\mathcal{D}_m$, batch size $b_s$}
    \KwOut{Network parameters $\theta$, soft label predictions $\hat{y}$}
    \While{not finished}{
        $\{x,\tilde{y}\}$ $\leftarrow$ GetBatch($\mathcal{D}_n$,$b_s$)\;
        $\{x_m,y_m\}$ $\leftarrow$ GetBatch($\mathcal{D}_m$,$b_s$)\;
        Update $\hat{y}$ by \autoref{eq:thetahat} and \autoref{eq:ymeta}\;
        Update $\theta$ by \autoref{eq:train}\;
     }
     \caption{Meta Label Noise Cleaner}
     \label{algo}
\end{algorithm}

\subsection{Formulation of $\hat{y}$}

In MSLG, label distribution $y^d$ is maintained for all training samples $x_i$. Following \cite{yi2019probabilistic}, we initialized $y^d$ using noisy labels $\tilde{y}$ with the following formula

\begin{equation}
    y^d = K\tilde{y}
\end{equation}

where $K$ is a large constant. Then softmax is applied to get normalized soft labels

\begin{equation}
    \hat{y} = softmax(y^d)
\end{equation}

This setup provides unconstrained learning for $y^d$ while producing valid soft labels $\hat{y}$ all the time.

\subsection{Loss Functions} \label{lossfunctions}

\subsubsection{\textbf{Classification Loss}}
Inspired from \cite{yi2019probabilistic}, we defined our meta loss as KL-divergence loss with a slight trick. KL-divergence is formulated as follows:

\begin{equation}
    KL(P||Q) = \sum_{x \in X} P(x) \log\left(\dfrac{P(x)}{Q(x)}\right)
\end{equation}

KL-divergence loss is asymmetric, which means 

\begin{equation}
    KL(Q||P) \neq KL(P||Q)
\end{equation}

Therefore we have two different configurations. First options is:

\begin{equation}
    \begin{split}
        \mathcal{L}_{c,1} =  \dfrac{1}{N} \sum_{i=1}^{N} KL(\hat{y}_i||f_\theta(x_i)), where \\
        KL(\hat{y}_i||f_\theta(x_i)) =  \sum_{j=1}^{C} \hat{y}_i^j \log \left(\dfrac{\hat{y}_i^j}{f_\theta^j(x_i)}\right)
    \end{split}
\end{equation}

which produces the following gradients

\begin{equation}
    \dfrac{\partial \mathcal{L}_{c,1}}{\partial f^j_\theta(x_i)} = - \dfrac{\hat{y}_i^j}{f_\theta^j(x_i)}
    \label{eq:grad1_f}
\end{equation}

Second possible configuration of loss function is

\begin{equation}
    \begin{split}
        \mathcal{L}_{c,2} =  \dfrac{1}{N} \sum_{i=1}^{N} KL(f_\theta(x_i)||\hat{y}_i), where \\
        KL(f_\theta(x_i)||\hat{y}_i) =  \sum_{j=1}^{C} f_\theta^j(x_i) \log \left(\dfrac{f_\theta^j(x_i)}{\hat{y}_i^j}\right)
    \end{split}
\end{equation}

which produces the following gradients

\begin{equation}
    \dfrac{\partial \mathcal{L}_{c,2}}{\partial f^j_\theta(x_i)} = 1 + \log \left(\dfrac{f_\theta^j(x_i)}{\hat{y}_i^j}\right)
    \label{eq:grad2_f}
\end{equation}

Lets assume a classification task where true label is 2 ($y_i^2=1$) but noisy label is given as 5 ($\tilde{y}_i^5=1$). Now we consider two updates on $\hat{y}_i^2$ and $\hat{y}_i^5$

\begin{itemize}
    \item \textbf{Case $\boldsymbol{\hat{y}_i^{j=2}}$:} $\hat{y}_i^2$ is initially very small, therefore $f^2_\theta(x_i) \gg \hat{y}_i^2$. In that case $\mathcal{L}_{c,1}$ will produce a very small gradients \eqref{eq:grad1_f} while $\mathcal{L}_{c,2}$ will produce a medium positive gradient \eqref{eq:grad2_f} as desired.
    \item \textbf{Case $\boldsymbol{\hat{y}_i^{j=5}}$:} $\hat{y}_i^5$ initially has peak value; however, due to internal robustness of network we expect  $f^5_\theta(x_i) \ll \hat{y}_i^5$. In that case $\mathcal{L}_{c,1}$ produce large negative gradient \eqref{eq:grad1_f} while $\mathcal{L}_{c,2}$ produce medium negative gradients \eqref{eq:grad2_f}.
\end{itemize}

As a result we can conclude the following statement: $\mathcal{L}_{c,1}$ focuses on learning from $y_i^5$ (noisy label) while $\mathcal{L}_{c,2}$ focuses on positive learning from $y_i^2$ (correct label) and negative learning from $y_i^5$ (noisy label). Therefore we believe $\mathcal{L}_{c,2}$ is better choice for our learning objective. 

Even though our loss formulation may be seen same as \cite{yi2019probabilistic}, its usage it totally different. While \cite{yi2019probabilistic} uses $\nabla_{\hat{y}} \mathcal{L}_c$ to update $\hat{y}$, we use second order derivative of $\nabla_{f_\theta} \mathcal{L}_c$ as a meta-objective which is further explained in \autoref{metaobjective}.

\subsubsection{\textbf{Entropy Loss}}
Inspired from \cite{tanaka2018joint}, we defined additional entropy loss as a regularization term. This extra loss forces network predictions to peak only at one class and zero out on others. This property is useful to prevent learning curve to halt since network predictions are forced to be different than estimated soft labels $\hat{y}$. Entropy loss is defined as follows

\begin{equation}
    \mathcal{L}_e(f_\theta(x)) = - \dfrac{1}{N} \sum_{i=1}^{N} \sum_{j=1}^{C} f^j_\theta(x_i) \log(f^j_\theta(x_i))
\end{equation}

\subsection{Meta Objective} \label{metaobjective}
From \autoref{eq:ymeta}, update term for $\hat{y}$ is as follows

\begin{equation}
    \beta \dfrac{1}{M} \sum_{i=1}^{M} \nabla_{\hat{y}} L^{m}_i(\hat{\theta}(\hat{y}))\Biggr|_{\hat{y}^{(t)}}
\end{equation}
\begin{equation}
    = \beta \dfrac{1}{M} \sum_{i=1}^{M} \frac{\partial L^{m}_i(\hat{\theta}(\hat{y}))}{\partial \hat{\theta}(\hat{y})} 
    \frac{\partial \hat{\theta}(\hat{y})}{\partial \hat{y}} \Biggr|_{\hat{y}^{(t)}}
\end{equation}

Using \autoref{eq:thetahat} for $\hat{\theta}(\hat{y})$ and replacing $\hat{\theta}(\hat{y})$ with $\hat{\theta}$ for the ease of notation we get

\begin{equation}
    = -\dfrac{\alpha\beta}{MN} \sum_{i=1}^{M} \frac{\partial L^{m}_i(\hat{\theta})}{\partial \hat{\theta}} 
    \frac{\partial}{\partial \hat{y}}\left(\sum_{j=1}^{N} \frac{\partial L^{n}_j(\theta,\hat{y})}{\partial \theta}\right)\Biggr|_{\hat{y}^{(t)}}  
\end{equation}

\begin{equation}
    = -\dfrac{\alpha\beta}{N} \sum_{j=1}^{N} \frac{\partial}{\partial \hat{y}} 
    \left(\dfrac{1}{M} \sum_{i=1}^{M} \frac{\partial L^{m}_i(\hat{\theta})}{\partial \hat{\theta}} \frac{\partial L^{n}_j(\theta,\hat{y})}{\partial \theta}\right)\Biggr|_{\hat{y}^{(t)}}  
\end{equation}

Let

\begin{equation}
    G_{ij}(\hat{y}) = \frac{\partial L^{m}_i(\hat{\theta})}{\partial \hat{\theta}} \frac{\partial L^{n}_j(\theta,\hat{y})}{\partial \theta}
\end{equation}

Then we can rewrite \autoref{eq:ymeta} as

\begin{equation}
    \hat{y}^{(t+1)} = \hat{y}^{(t)} + \dfrac{\alpha\beta}{N} \sum_{j=1}^{N} \frac{\partial}{\partial \hat{y}} 
    \left(\dfrac{1}{M} \sum_{i=1}^{M} G_{ij}(\hat{y})\right)\Biggr|_{\hat{y}^{(t)}}
\end{equation}

In this formulation $\dfrac{1}{M} \sum_{i=1}^{M} G_{ij}(\hat{y})$ represents the similarity between the gradient of the $j^{th}$ training sample subjected to $\theta$ and the mean gradient  computed over the batch of meta-data. Therefore, this will peak when gradients on a training sample and mean gradients over a mini-batch of meta samples are most similar. As a result, taking a gradient step subjected to $\hat{y}$ means finding the optimal label distribution so that produced gradients from training data are similar with gradients from meta-data.

\subsection{Overall Algorithm}
We propose a two-stage framework for our algorithm. In the first stage we train the base classifier with traditional SGD on noisy labels as warm-up training, and in the second stage we employ MSLG algorithm \autoref{algo}.

Warm-up training is advantageous for two reasons. Firstly, works show that in the presence of noisy labels, deep networks initially learn useful representations and overfit the noise only in the later stages \cite{krueger2017deep,arpit2017closer}. Therefore, we can still leverage useful information in initial epochs. Secondly, we update $\hat{y}$ according to predictions coming from the base network. Without any pre-training, random feedback coming from the base network would cause $\hat{y}$ to lead in the wrong direction. 

	\section{Experiments} \label{experiments}

\subsection{Datasets}
\subsubsection{\textbf{CIFAR10}}
CIFAR10 has 60k images for ten different classes. We separated 5k images for the test set and another 5k for meta-set. The remaining 50k images are corrupted with synthetic label noise.

For synthetic noise, we used two types of noises; uniform noise and feature-dependent noise. For uniform noise, labels are flipped to any other class uniformly with the given error probability. For feature-dependent noise, we followed \cite{algan2020label}, in which a pre-trained network is used as feature extractor to map images to feature domain. Then samples that are closest to decision boundaries are flipped to its counter class. This noise checks the image features and finds the most ambiguous samples and flip their label to the class of most resemblance.

\subsubsection{\textbf{Clothing1M}}
Clothing1M is a large-scale dataset with one million images collected from the web \cite{xiao2015learning}. It has images of clothings from fourteen classes. Labels are constructed from surrounding texts of images and are estimated to have a noise rate of around 40\%. There exists 50k, 14k and 10k additional verified images for training, validation and test set. We used 14k validation set as meta-data and 10k test samples to evaluate the classifier's final performance. In order to have a fair evaluation, we did not use 50k clean training samples at all. 

\subsubsection{\textbf{Food101N}}
Food101N is an image dataset containing about 310k images of food recipes classified in 101 classes \cite{lee2018cleannet}. It shares the same classes with Food101 dataset but has much more noisy labels, which is estimated to be around 20\%. It has 53k verified training and 5k verified test images. We used 15k samples from verified training samples as meta-dataset.

\subsection{Implementation Details}
For all datasets we used SGD optimizer with 0.9 momentum and $10^{-4}$ weight decay. We set $K=10$ for all our experiments.

\subsubsection{\textbf{CIFAR10}}
We use an eight-layer convolutional neural network with six convolutional layers and two fully connected layers. The batch size is set as 128. $\lambda$ is initialized as $10^{-2}$ and set to $10^{-3}$ and $10^{-4}$ at $40^{th}$ and $80^{th}$ epochs. Other parameters are set as \autoref{cifar10params}. Total training consists of 120 epochs, in which the first 44 epochs are warm-up and the remaining epochs are MSLG training. For data augmentation, we used random vertical and horizontal flip. Moreover, we pad images 4 pixels from each side and random crop 32x32 pixels.

\subsubsection{\textbf{Clothing1M}}
In order to have a fair comparison with the works from the literature, we followed the widely used setup of ResNet-50 architecture pre-trained on ImageNet. Batch size is set to 32. $\lambda$ is set to $10^{-3}$ for the first 5 epochs and to $10^{-4}$ for the second 5 epochs. Total training consists of 10 epochs, in which the first epoch is warm-up and the rest is MSLG training. Other parameters are set as follow; $\alpha=0.1, \beta=100$. All images are resized to 256x256, and then central 224x224 pixels are taken.

\subsubsection{\textbf{Food101N}} We used the same setup and parameter set from Clothing1M. Only difference are at the following parameters; $\lambda=0.5, \beta=1500$

\subsection{Experiments on CIFAR10}
We tested our algorithm with CIFAR10 under varying level of noise ratios for different types of noises. Since we manually add synthetic noise, we have complete knowledge over the noise. Therefore, we feed the exact noise transition matrix to forward loss method \cite{patrini2017making} for baseline comparison, which is not possible in real-world datasets. Results are presented in \autoref{tbl:cifar10}. 

As can be seen, we beat all baselines with a large margin for all noise rates of the feature-dependent noise. We manage to get around 75\% accuracy even under the extreme case of 80\% noise. For uniform noise, our algorithm falls shortly behind the best model, but still manages to get comparable results. This can be explained as follow. For uniform noise, noisy samples are totally unrelated to features and true label of the data. Due to the internal robustness of the network, this would result in $f^j_\theta(x_i) \ll \tilde{y}^j_i$ for noisy class $j$. This would result in large negative gradients \autoref{eq:grad2_f}. In the case of feature-dependent noise, noisy labels are related to real label distribution. As a result, network prediction and noisy label is more similar $f^j_\theta(x_i) < \tilde{y}^j_i$ for noisy class $j$. This would result in smaller gradients \autoref{eq:grad2_f}. Therefore, when noise is random, gradients due to noisy class may overcome the gradients due to true class, which presents a more challenging framework for stabilization of robust learning. Therefore, our proposed framework is slightly less robust to random noise.

\begin{figure}
	\includegraphics[width=0.5\textwidth]{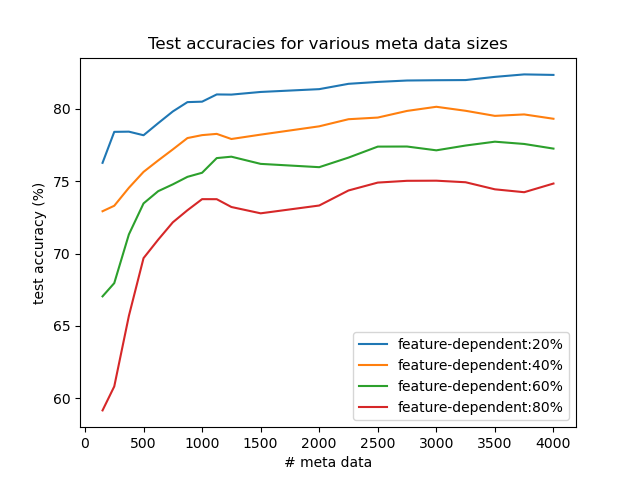}
	\caption{Test accuracies for different numbers of meta-data.}
	\label{fig:metasizes}
\end{figure}

We test our algorithm with changing number of meta-data samples. As expected, our accuracy increases with increasing number of meta-data. However, as can be seen from \autoref{fig:metasizes}, only small amount of meta-data is required for our framework. For CIFAR10 dataset with 50k noisy training samples, 1k meta-data (2\% of training data) achieves approximately the top result. Moreover, as presented in \autoref{fig:metasizes}, number of required meta-data is independent of the noise ratio.

\begin{table*} \centering
    \begin{tabular}{l|c|cccc|cccc}
        \noalign{\vskip 0.3cm} 
        \hline 
        noise type                               &                         & \multicolumn{4}{c|}{uniform}                                                                          & \multicolumn{4}{c}{feature-dependent}                             \\ \hline
        noise ratio (\%)                         & 0                       & 20                      & 40                      & 60                      & 80                      & 20                      & 40                      & 60                      & 80                      \\ \hline
        Cross Entropy                            & 88.13          & 82.69$\pm$0.44          & 76.84$\pm$0.31          & 66.46$\pm$0.60          & 38.04$\pm$0.92          & 81.21$\pm$0.11          & 71.46$\pm$0.22          & 69.19$\pm$0.61          & 23.89$\pm$0.38          \\
        Generalized-CE\cite{zhang2018generalized}& 86.34          & 84.62$\pm$0.25          & 81.98$\pm$0.19          & \textbf{74.48$\pm$0.43} & 44.36$\pm$0.63          & 81.21$\pm$0.22          & 71.70$\pm$0.54          & 66.56$\pm$0.36          & 10.93$\pm$0.68          \\
        Bootstrap \cite{reed2014training}        & 88.24          & 82.51$\pm$0.21          & 76.97$\pm$0.23          & 66.13$\pm$0.36          & 38.41$\pm$1.65          & 81.24$\pm$0.15          & 71.63$\pm$0.42          & 69.74$\pm$0.32          & 23.25$\pm$0.20          \\
        Co-Teaching\cite{han2018co}              & 84.72          & \textbf{85.96$\pm$0.09} & 80.24$\pm$0.12          & 70.38$\pm$0.15          & 41.22$\pm$2.07          & 81.19$\pm$0.27          & 72.47$\pm$0.19          & 67.67$\pm$0.79          & 18.66$\pm$0.17          \\
        Forward Loss\cite{patrini2017making}     & 85.36          & 83.31$\pm$0.45          & 80.25$\pm$0.11          & 71.34$\pm$0.30          & 28.77$\pm$0.19          & 77.60$\pm$0.34          & 69.21$\pm$0.13          & 39.23$\pm$1.98          & 11.61$\pm$0.48          \\
        Symmetric-CE\cite{wang2019symmetric}     & 84.66          & 82.72$\pm$0.32          & 79.79$\pm$0.44          & 74.09$\pm$0.32          & 54.56$\pm$0.48          & 76.21$\pm$0.16          & 67.76$\pm$3.11          & fail                    & fail                    \\
        Joint Opt.\cite{tanaka2018joint}         & \textbf{88.70} & 83.74$\pm$0.29          & 78.75$\pm$0.38          & 68.17$\pm$0.62          & 39.22$\pm$1.04          & 81.61$\pm$0.16          & 74.03$\pm$0.13          & 72.15$\pm$0.32          & 44.15$\pm$0.30          \\
        MLNT\cite{junnan2018learning}            & 87.50          & 83.20$\pm$0.21          & 78.14$\pm$0.13          & 66.34$\pm$0.57          & 40.80$\pm$1.31          & 82.46$\pm$0.24          & 72.52$\pm$0.38          & 70.12$\pm$0.51          & 41.12$\pm$0.82          \\ 
        PENCIL\cite{yi2019probabilistic}         & 87.04          & 83.34$\pm$0.28          & 79.27$\pm$0.29          & 71.41$\pm$0.14          & 46.57$\pm$0.37          & 81.62$\pm$0.12          & 75.08$\pm$0.28          & 69.24$\pm$0.39          & 10.71$\pm$0.10          \\
        Meta-Weight\cite{shu2019meta}            & 88.38          & 84.12$\pm$0.29          & \textbf{80.68$\pm$0.16} & 71.78$\pm$0.38          & 46.71$\pm$2.18          & 81.06$\pm$0.22          & 71.50$\pm$0.27          & 67.50$\pm$0.45          & 21.28$\pm$1.01          \\ \hline
        \textbf{MSLG}                            & 87.55          & 83.48$\pm$0.11          & 79.82$\pm$0.24          & 72.92$\pm$0.18          & \textbf{56.26$\pm$0.20} & \textbf{82.62$\pm$0.05} & \textbf{79.30$\pm$0.23} & \textbf{77.33$\pm$0.16} & \textbf{74.87$\pm$0.11} \\ \hline 
    \end{tabular}
    \caption{Test accuracy percentages for CIFAR10 dataset with varying level of uniform and feature-dependent noise. Results are averaged over 4 runs.}
    \label{tbl:cifar10}
\end{table*} 

\begin{table}[] \centering
    \begin{tabular}{c|c|c}
        \hline
        noise ratio & uniform    & feature-dependent \\ \hline
        20\% & $\alpha=0.5, \beta=4000$ & $\alpha=0.5, \beta=4000$\\
        40\% & $\alpha=0.5, \beta=4000$ & $\alpha=0.5, \beta=4000$\\  
        60\% & $\alpha=0.5, \beta=2000$ & $\alpha=0.5, \beta=4000$\\
        80\% & $\alpha=0.5, \beta=400$ & $\alpha=0.5, \beta=4000$\\ \hline
    \end{tabular}
    \caption{Hyper-parameters for CIFAR10 experiments.}
    \label{cifar10params}
\end{table}

\subsection{Experiments on Clothing1M}
In order to show effectiveness of our algorithm, we tested with real-world noisy label dataset Clothing1M. For forward loss method \cite{patrini2017making}, we used samples with both verified and noisy labels to construct noise transition matrix. Results are presented in \autoref{results_clothing1m}, where we managed to outperformed all baselines. Our proposed algorithm achieves 76.02\% test accuracy, which is 2.3\% higher than state of the art.

\begin{table}[]
    \begin{tabular}{l|l|l|l}
        \hline
        \multicolumn{1}{c|}{method} & \multicolumn{1}{c|}{accuracy}  & \multicolumn{1}{c|}{method} & \multicolumn{1}{c}{accuracy}   \\ \hline
        Cross Entropy                             & 69.93        & Symmetric CE \cite{wang2019symmetric}    & 71.02                 \\
        Generalized CE \cite{zhang2018generalized}& 67.85        & Joint Optimization \cite{tanaka2018joint}& 72.16                 \\  
        Bootstrap \cite{reed2014training}         & 69.35        & MLNT \cite{junnan2018learning}           & 73.47                 \\
        Co-Teaching \cite{han2018co}              & 69.63        & PENCIL \cite{yi2019probabilistic}        & 73.49                 \\ 
        Forward Loss \cite{patrini2017making}     & 70.94        & Meta-Weight Net \cite{shu2019meta}       & 73.72                 \\ \hline
        \multicolumn{4}{c}{\textbf{MSLG: 76.02}} \\ \hline
    \end{tabular}
    \caption{Test accuracy percentages on Clothing1M dataset. Results for cross-entropy, forward loss \cite{patrini2017making}, generalized cross-entropy \cite{zhang2018generalized}, bootstrap \cite{reed2014training}, co-teaching \cite{han2018co} are obtained from our own implementations. Rest of the results are taken from the corresponding paper.}
    \label{results_clothing1m}
\end{table}

\subsection{Experiments on Food101N}
In order to further test our algorithm under real-world noisy label data, we conduct tests on Food101N dataset too. Since none of the baselines provided results on Food101N dataset, all results are taken from our own implementations. There are excessively large number of classes (101) in the dataset, hence some methods fail to succeed. For example, methods depending on noise transition matrix fail since the matrix becomes intractably large. We only present results of methods which had fair performance. This dataset has a much smaller noise ratio (20\%), as a result all algorithms results around similar accuracies with straight forward training with cross-entropy loss. Therefore, there are no big gaps among top accuracies, but still as presented in \autoref{results_food101N}, our proposed framework manages to get best accuracy in this dataset too.

\begin{table}[]
    \begin{tabular}{l|l|l|l}
        \hline
        \multicolumn{1}{c|}{method} & \multicolumn{1}{c|}{accuracy}  & \multicolumn{1}{c|}{method} & \multicolumn{1}{c}{accuracy} \\ \hline
        Cross Entropy                             & 77.51        & Joint Optimization \cite{tanaka2018joint}& 76.12               \\
        Generalized CE \cite{zhang2018generalized}& 71.60        & PENCIL \cite{yi2019probabilistic}        & 78.26               \\  
        Bootstrap \cite{reed2014training}         & 78.03        & Meta-Weight Net \cite{shu2019meta}       & 76.12               \\
        Co-Teaching \cite{han2018co}              & 78.95        & \textbf{MSLG}                            & \textbf{79.06}      \\ \hline
    \end{tabular}
    \caption{Test accuracy percentages for Food101N dataset. All values in the table are obtained from our own implementations.}
    \label{results_food101N}
\end{table}

	\section{Conclusion} \label{conclusion}
We proposed a meta soft label generation framework to train deep networks in the presence of noisy labels. Our approach seeks optimal label distribution subjected to meta-objective, which is to minimize the loss of the small meta-dataset. Afterwards, the network is trained on these predicted soft labels. These two stages are repeated consecutively during the training process. Our proposed solution is model agnostic and can easily be adapted to any system at hand.

We conduct extensive experiments on CIFAR10 dataset with various levels of synthetic noise for both uniform and feature-dependent noise scenarios. Experiments showed that our framework outperforms all baselines with a large margin, especially in the case of structured noises. For further evaluation, we conduct tests on real-world noisy label dataset Clothing1M, where we beat state of the art with more than 2\%. Moreover, for further evaluation, we tested our algorithm on another real-world noisy labeled dataset Food101N, where we outperformed all other baselines.

	\bibliographystyle{IEEEtran}
	\bibliography{paperrefs}

\end{document}